\documentclass[11pt,a4paper]{article}
\usepackage[hyperref]{naaclhlt2019_mine}
\usepackage{times}
\usepackage{latexsym}
\usepackage{tabulary}
\usepackage{amssymb}
\usepackage{amsmath}
\usepackage{graphicx}
\usepackage{url}

\aclfinalcopy 
\def\aclpaperid{jgkimi} 

\title{Decay-Function-Free Time-Aware Attention to Context and Speaker Indicator for Spoken Language Understanding}

\author{Jonggu Kim \\
  Computer Science and Engineering, \\
  Pohang University of Science and \\ Technology (POSTECH) \\
  Pohang, Republic of Korea \\
  {\tt jgkimi@postech.ac.kr} \\\And
  Jong-Hyeok Lee \\
  Computer Science and Engineering, \\
  Pohang University of Science and \\ Technology (POSTECH) \\
  Pohang, Republic of Korea \\
  {\tt jhlee@postech.ac.kr} \\}

\date{}

\begin{document}
\maketitle
\begin{abstract}
  To capture salient contextual information for spoken language understanding (SLU) of a dialogue, we propose time-aware models that automatically learn the latent time-decay function of the history without a manual time-decay function. We also propose a method to identify and label the current speaker to improve the SLU accuracy. In experiments on the benchmark dataset used in Dialog State Tracking Challenge 4, the proposed models achieved significantly higher F1 scores than the state-of-the-art contextual models. Finally, we analyze the effectiveness of the introduced models in detail. The analysis demonstrates that the proposed methods were effective to improve SLU accuracy individually.
\end{abstract}

\begin{figure*}[t]
  \centering
  \noindent
  \includegraphics[width=\linewidth]{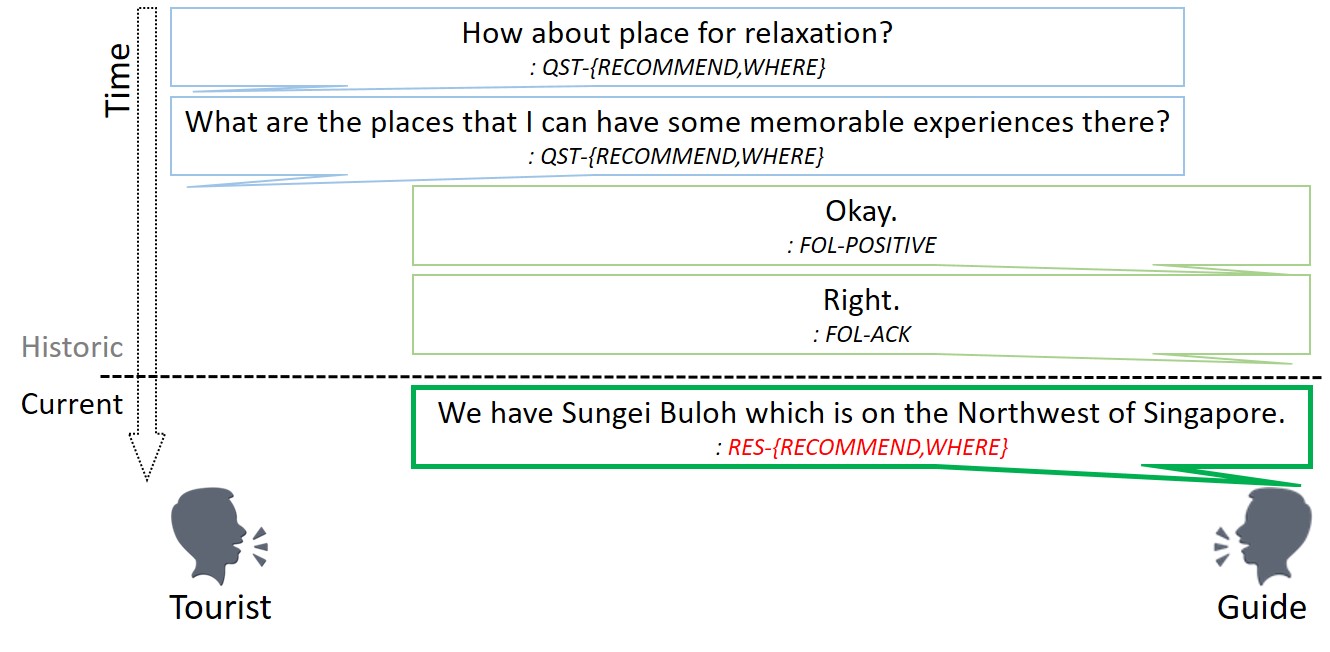}
  \caption{An example of utterances with their semantic labels (speech acts combined with associated attributes) from DSTC 4. The semantic labels are italicized.}
  \label{fig:conversation}
\end{figure*}

\section{Introduction}
  Spoken language understanding (SLU) is a component that understands the user's utterance of a dialogue system. Given an utterance, SLU generates a structured meaning representation of the utterance; i.e., a semantic frame. SLU can be decomposed into several subtasks such as domain identification, intent prediction and slot filling; these subtasks can be jointly assigned using a single model \cite{Hakkani-Tur:INTERSPEECH16,Liu:INTERSPEECH16,Chen:SLT16}. The accuracy of SLU is important for the dialogue system to generate an appropriate response to a user. 

To improve the accuracy of SLU, much work has used contextual information of dialogues to alleviate the ambiguity of recognition of the given utterance. In SLU, selecting important history information is crucial, and it directly influences the improvement of SLU accuracy. To summarize this history, content-aware models \cite{Chen:INTERSPEECH16,Kim:ASRU17} similar to attention models in machine translation \cite{Bahdanau:arxiv14} have been proposed. However, content-aware models are likely to select the wrong history when the histories are similar in content. To alleviate this problem, time-aware models \cite{Chen:ASRU17,Su:NAACL18,Su:arxiv18} which pay attention to recent previous utterances by using the temporal distance between a previous utterance and a current utterance are being considered; the models are based on mathematical formulas, time-decay functions, which are formulated by human, and decomposed into trainable parameters.

However, the previous time-aware models may not be sufficiently accurate. In the models, either a single time-decay function is used or a limited number of time-decay functions are linearly combined; these manual functions may not be sufficiently flexible to learn an optimal time-decay function.




In this paper, we propose flexible and effective time-aware attention models to improve SLU accuracy. The proposed models do not need any manual time-decay function, but learn a time-decay tendency directly by introducing a trainable distance vector, and therefore have good SLU accuracy. The proposed models do not use long short-term memory (LSTM) to summarize histories, and therefore use fewer parameters than previous time-aware models. We also propose current-speaker modeling by using a speaker indicator that identifies the current speaker.

To the best of our knowledge, this is the first method that shows improvement by considering the identity of the current speaker. This information may be helpful for modeling multi-party conversations in addition to human-human conversations.

Prediction of the semantic label of the current utterance even using a conventional time-aware model can be difficult. (Figure \ref{fig:conversation}). The nearest utterance is ``Right.'', but it is not the most relevant utterance to the current utterance; the most relevant utterance is ``What are the places that I can have some memorable experiences there?''. If we do not know the current speaker is \textit{Guide}, we cannot easily assess the relative importance of the nearest histories of the two speakers. We believe that the proposed `speaker indicator' can help our model to identify such information.


In experiments on the Dialog State Tracking Challenge 4 (DSTC 4) dataset, the proposed models achieved significantly higher accuracy than the state-of-the-art contextual models for SLU. Also, we examine how the proposed methods affect the SLU accuracy in detail. This result shows that the proposed methods were effective to improve SLU accuracy individually. Our contributions are as follows:

\begin{itemize}
\item We propose a decay-function-free time-aware attention model that automatically learn the latent time-decay function of the history without a manual time-decay function. The proposed model achieves a new state-of-the-art F1 score.
\item We propose a current-speaker modeling method that uses a speaker indicator to identify the current speaker. We present how to incorporate speaker indicator in the proposed attention model for further improvement of SLU accuracy.
\item We propose a model that is aware of content as well as time, which also achieved a higher F1 score than the state-of-the-art contextual models.
\item We analyze the effectiveness of proposed methods in detail.
\end{itemize}

Our source code to reproduce the experimental results is available at \url{https://github.com/jgkimi/Decay-Function-Free-Time-Aware}.

\begin{figure*}[t]
  \centering
  \noindent
  \includegraphics[width=\linewidth]{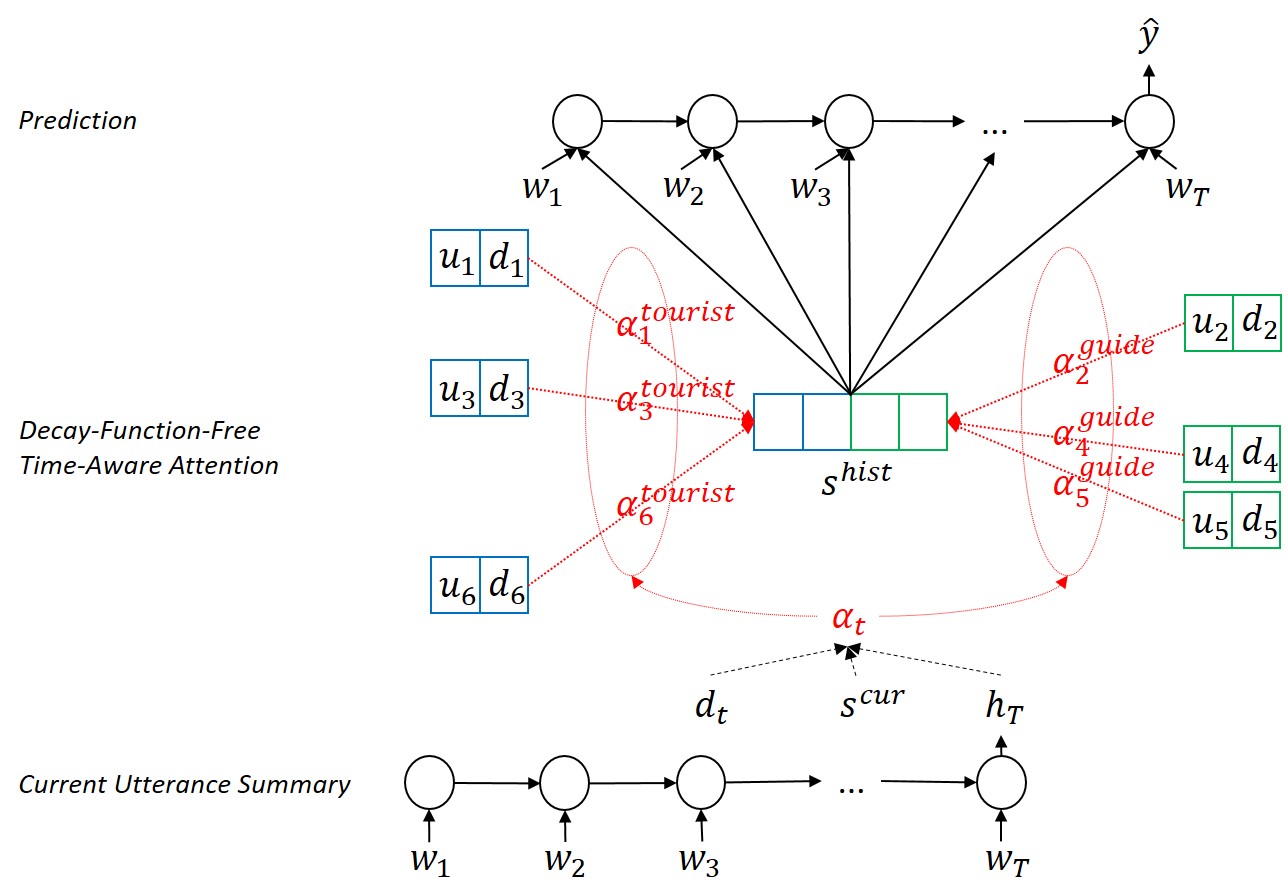}
  \caption{Overall architecture of the decay-function-free time-aware attention with speaker indicator (role-level). \(w_1,..., w_T\) are word vectors of the current utterance, \(d_t\) is the \(t^{th}\) distance vector, \(u_t\) is the \(t^{th}\) utterance vector, \(s^{cur}\) is the current speaker indicator, \(h_T\) is the current utterance summary vector and \(\alpha_t\) is the importance of the \(t^{th}\) historic utterance. For simplicity, we represent bidirectional LSTM layers as unidirectional LSTM layers.}
  \label{fig:architecture}
\end{figure*}

\section{Related Work}
Joint semantic frame parsing has the goal of learning intent prediction and slot filling jointly. By joint learning, the model learns their shared features, and this ability is expected to improve the accuracy on both tasks. A model based on bidirectional LSTM for joint semantic frame parsing \cite{Hakkani-Tur:INTERSPEECH16} is trained on the two tasks in sequence, by adding an intent label to the output of the final time-step of LSTM. Similarly, an attention-based LSTM predicts slot tags for each time-step, then feeds the hidden vectors and their soft-aligned vectors to a fully-connected layer for intent prediction \cite{Liu:INTERSPEECH16}. Knowledge-guided joint semantic frame parsing \cite{Chen:SLT16} incorporates syntax or semantics-level parsing information into a model by using a recurrent neural network (RNN) for joint semantic frame parsing.

Other research on SLU uses context information. A model based on support vector machine and a hidden Markov model uses contextual information to show the importance of contextual information in SLU tasks, intent prediction and slot detection \cite{Bhargava:ICASSP13}. RNN-based models can exploit context to classify domains \cite{Xu:ICASSP14}, and have been combined with previously-estimated intent and slot labels to predict domain and intent \cite{Shi:ICASSP15}. A memory network that contains historic utterance vectors encoded by RNN has been used to select the most relevant history vector by multiplicative soft-attention \cite{Chen:INTERSPEECH16}; the selected vector is fed to an RNN-based slot tagger as context information.

A memory network can be regarded as use of content-based similarity between the current utterance and previous utterances. A memory network can be separated to capture historic utterances for each speaker independently \cite{Kim:ASRU17}, and a contextual model can use different LSTM layers to separately encode a history summary for each speaker \cite{Chi:IJCNLP17}. For another task, addressee and response selection in multi-party conversations, a distinct RNN-based encoder for each speaker-role (sender, addressee, or observer) has been used to generate distinct history summaries \cite{Zhang:AAAI18}.

Recent work on contextual SLU has introduced time information of contexts into models because content-based attention may cause a wrong choice that introduce noises. The reciprocal of temporal distance between a current utterance and contexts can be used as a time-decay function, and the function can be decomposed into trainable parameters \cite{Chen:ASRU17}. Similarly, a universal time-aware attention model \cite{Su:NAACL18} has been proposed; it is a trainable linear combination of three distinct (convex, linear and concave) time-decay functions . An extension of this model is a context-sensitive time-decay attention \cite{Su:arxiv18} that generates its parameters from the current utterance by using a fully-connected layer so that the content information of the current utterance is also considered in the attention.

\section{Proposed Model}\label{sec:proposed_model}
We propose a time-aware model that includes a speaker indicator (Figure \ref{fig:architecture}). In addition, we propose a content-and-time-aware model that includes a speaker indicator. The models are trained in an end-to-end way, in which every model parameter is automatically learned based on a downstream SLU task. The objective of the proposed models are to optimize the conditional probability of labels of SLU, given the current utterance \(p(\hat{y}|x)\), by minimizing the cross-entropy loss.

The following description of the proposed model considers three steps: current utterance summary, context summary, and prediction.

\subsection{Current Utterance Summary}
To select salient parts of contextual histories, the current utterance is used. To summarize a current utterance matrix \(U\) that consists of words \(w_i\) as a vector (i.e., \(U = \{w_1\) , \(w_2\),..., \(w_T\)\}), \(U\) is fed to bidirectional LSTMs, and the final hidden vector \(h_T \in \mathbb{R}^{dim}\) is taken as a current utterance summary.

\subsection{Decay-Function-Free Time-Aware Attention}
In this subsection, we introduce a decay-function-free time-aware attention model. To summarize contexts, we use a time difference (distance) between a historic utterance and the current utterance; this distance represents the interval between the historic utterance and the current utterance. We use the distance of the \(t^{th}\) history from the current utterance as an index to select a dense vector from a distance-embedding matrix \(D \in \mathbb{R}^{dim \times |D|}\), then use the vector as the \(t^{th}\) distance vector \(d_t\).

To compute the importance \(\alpha_t\) of the \(t^{th}\) history, both in the sentence-level attention and in the role-level attention, our time-aware attention uses the current utterance summary \(h_T\) and the history distance \(d_t\) simultaneously and additively:
\begin{flalign}
  & \alpha_t = w_{att}^T \sigma \big(h_T + d_t + b_{att}\big) ,
  \label{eq:cal_alpha}
\end{flalign}
where \(w_{att}^T\) is the transpose of a trainable weight vector for the attention, \(b_{att}\) is a trainable bias vector for the attention, and \(\sigma\) is the hyperbolic tangent function.

Computing a time-aware context summary vector \(s_{time}^{hist}\) depends on whether the role-level or sentence-level attention is considered. For the role-level attention, we use the softmax operation applied to all \(\alpha_t\) of the same speaker, either a guide or a tourist, to obtain a role-level probabilistic importance \(\alpha_t^{role}\) of \(t^{th}\) history. We then multiply \(\alpha_t^{role}\) by \(t^{th}\) history vector, which is a concatenation of the corresponding intent-dense vector \(u_t\) and the distance vector \(d_t\). We use the element-wise sum of the vectors of the same speaker to construct two summary vectors \(s_{time}^{guide}\) and \(s_{time}^{tourist}\). Finally, \(s_{time}^{guide}\) and \(s_{time}^{tourist}\) are concatenated to form a time-aware history summary vector \(s_{time}^{hist}\) as:

\begin{flalign}
  & \alpha_t^{role} = softmax_{role}(\alpha_t), \\
  & s_{time}^{role} = \displaystyle\sum_{t} \alpha_{t}^{role}(u_t\oplus d_t) , \\
  & s_{time}^{hist} = s_{time}^{guide} \oplus s_{time}^{tourist},
\end{flalign}
where \(\oplus\) represents a concatenation operation.

For the sentence-level attention to obtain a sentence-level probabilistic importance \(\alpha_t^{sent}\) of \(t^{th}\) history, we use the softmax operation applied to all \(\alpha_t\) regardless of the speaker, then multiply \(\alpha_t^{sent}\) by the \(t^{th}\) history vector, which is a concatenation of the corresponding intent dense vector \(u_t\) and the distance vector \(d_t\). We use the element-wise sum of the vectors to construct a time-aware summary vector \(s_{time}^{hist}\) as:
\begin{flalign}
  & \alpha_t^{sent} = softmax_{sent}(\alpha_t), \\
  & s_{time}^{hist} = \displaystyle\sum_{t} \alpha_{t}^{sent}(u_t\oplus d_t).
\end{flalign}

Then, \(s_{time}^{hist}\) is used as a context summary \(s^{hist}\) in the prediction step.

\subsection{Decay-Function-Free Content-and-Time-Aware Attention}
Although a time-aware attention model is powerful by itself, content can be considered at the same time to improve accuracy. We propose another contextual attention model that is aware of content, in addition to time. This model is called content-and-time-aware attention. The model uses an importance value \(\beta_t\) for the \(t^{th}\) history. To compute \(\beta_t\), we uses the trainable parameters \(w_{att}\) and \(b_{att}\) of the time attention as:
\begin{flalign}
  & \beta_t = w_{att}^T \sigma \big(h_T + u_t + b_{att}\big) ,
  \label{eq:cal_beta}
\end{flalign}
where \(u_t\) is the intent dense vector of \(t^{th}\) history, and \(\sigma\) is the hyperbolic tangent function.

Then, \(\beta_t\) is used in the same way as \(\alpha_t\), but independently. \(s_{time}^{hist}\) is computed as in the previous subsection, \(\beta_t\) is used to compute \(s_{cont}^{hist}\) for the role-level attention as:
\begin{flalign}
  & \beta_t^{role} = softmax_{role}(\beta_t), \\
  & s_{cont}^{role} = \displaystyle\sum_{t} \beta_{t}^{role}(u_t\oplus d_t) , \\
  & s_{cont}^{hist} = s_{cont}^{guide} \oplus s_{cont}^{tourist}.
\end{flalign}

To compute \(s_{cont}^{hist}\) for the sentence-level attention, \(\beta_t\) is used as:
\begin{flalign}
  & \beta_t^{sent} = softmax_{sent}(\beta_t), \\
  & s_{cont}^{hist} = \displaystyle\sum_{t} \beta_{t}^{sent}(u_t\oplus d_t).
\end{flalign}

Finally, the time-aware history summary \(s_{time}^{hist}\) and the content-aware history summary \(s_{cont}^{hist}\) are concatenated to generate a history summary \(s^{hist}\) regardless of the attention level:
\begin{flalign}
  & s^{hist} = s_{time}^{hist} \oplus s_{cont}^{hist}.
\end{flalign}

\begin{table*}[t]
  \centering
  \begin{tabulary}{5.0\textwidth}{p{10cm}|c|c}
    \hline
    &\multicolumn{2}{c}{\textbf{F1 score}} \\
    \textbf{Model} & \textbf{Sent.-Level} & \textbf{Role-Level} \\
    \hline
    \hline
    DSTC 4 - Best & \multicolumn{2}{c}{\textit{61.4}} \\
    No Context & \multicolumn{2}{c}{65.06} \\
    LSTM-Used Context Summary without Attention &  \multicolumn{2}{c}{72.15} \\
    LSTM-Used Content-Aware Attention & 71.27 & 71.84\\ 
    Speaker Role Modeling \cite{Chi:IJCNLP17} & \textit{66.8} & \textit{70.1} \\
    Convex Time-Aware Attention \cite{Chen:ASRU17} & \textit{74.6} & \textit{74.2} \\
    Universal Time-Aware Attention \cite{Su:NAACL18} & 74.22 & 74.12\\
    Universal Content + Time Attention \cite{Su:NAACL18} & 74.40 & 74.33\\
    Context-Sensitive Time Attention \cite{Su:arxiv18} & 74.20 & 73.53\\
    \hline
    Decay-Function-Free Time-Aware Attention & 75.58\(^{**}\) & 75.58\(^{**}\) \\ 
    \hspace{0.5cm} with Speaker Indicator & 75.95\(^{**}\) & \textbf{76.56}\(^{**}\) \\
    Decay-Function-Free Content-and-Time-Aware Attention & 75.59\(^{**}\) & 75.30\(^{**}\) \\ 
    \hspace{0.5cm} with Speaker Indicator & 76.11\(^{**}\) & 76.14\(^{**}\) \\ 
    \hline
  \end{tabulary}
  \caption{SLU accuracy on DSTC 4. \(^{*}\): \(p < 0.05\); \(^{**}\), \(p < 0.01\) compared to all the baseline models. Italicized scores are reported in the references. Model names are described in the text.}
  \label{table:result}
\end{table*}

\subsection{Speaker Indicator}
Speaker indicator is a trainable vector \(s^{cur} \in \mathbb{R}^{dim}\) which indicates the identity of the current speaker; i.e., either a tourist or a guide in DSTC 4. An embedding lookup method is used after a speaker embedding matrix \(S \in \mathbb{R}^{dim \times |S|}\) is defined. The speaker embedding matrix is randomly initialized before the model is trained.

To use speaker indicator \(s^{cur}\) in the proposed attentions, Eq. \ref{eq:cal_alpha} is rewritten as:
\begin{flalign}
  & \alpha_t = w_{att}^T \sigma \big(h_T + d_t + s^{cur} + b_{att}\big),
\end{flalign}

and Eq. \ref{eq:cal_beta}  is rewritten as:
\begin{flalign}
  & \beta_t = w_{att}^T \sigma \big(h_T + u_t + s^{cur} + b_{att}\big).
\end{flalign}

\subsection{Prediction}
To predict the true label in spoken language understanding, our model consumes the current utterance \(U\) again. We use another bidirectional LSTM layer which is distinct from that of the current utterance summary. To prepare for \(t^{th}\) input \(v_t\) of the LSTM layer, we concatenate \(t^{th}\) word vector \(w_t\) of the current utterance \(U\) with the history summary vector \(s^{hist}\):

\begin{flalign}
v_t = w_t \oplus s^{hist}.
\end{flalign}

Then, we feed each \(v_t\) to the LSTM layer sequentially, and the final hidden vector of the LSTM layer is used as an input of a feed-forward layer to predict the true label \(\hat{y}\).

\section{Experiments}
To test the proposed models, we conducted language-understanding experiments on a dataset of human-human conversations.

\subsection{Dataset and Settings}
We conducted experiments on the DSTC 4 dataset which consists of 35 dialogue sessions on touristic information for Singapore; they were collected from Skype calls of three tour guides with 35 tourists. The 35 dialogue sessions total 21 h, and include 31,034 utterances and 273,580 words \cite{Kim:IWSDS16}. DSTC 4 is a suitable benchmark dataset for evaluation, because all of the dialogues have been manually transcribed and annotated with speech acts and semantic labels at each turn level. a semantic label consists of a speech act and associated attribute(s). The speaker information (guide and tourist) is also provided. Human-human dialogues contain rich and complex human behaviors and bring much difficulty to all tasks that are involved in SLU. We used the same training dataset, the same test dataset and the same validation set as in the DSTC 4 competition: 14 dialogues as the training dataset, 6 dialogues as the validation dataset, and 9 dialogues as the test dataset.

We used Adam \cite{Kingma:ICLR15} as the optimizer in training the model. We set the batch size to 256, and used pretrained 200-dimensional word embeddings GloVe \cite{Pennington:EMNLP14}. We applied 30 training epochs with early stopping. We set the size \(dim\) of every hidden layer to 128, and the context length to 7. We used the ground truth intents (semantic labels) to form an intent-dense vector like previous work. To evaluate SLU accuracy, we used the F1 score, which is the harmonic mean of precision and recall. To validate the significance of improvements, we used a one-tailed t-test. We ran each model ten times, and report their average scores.

As baseline models, we used the state-of-the-art contextual models, and most accurate participant of DSTC 4 (DSTC 4 - Best) \cite{Kim:IWSDS16}. For comparison with our models, we used the scores reported in the papers\footnote{\newcite{Su:NAACL18} and \newcite{Su:arxiv18} specified that they used different training/valid/test datasets that had been randomly selected from the whole DSTC 4 data with different rates for the experiments. Therefore, we do not use the reported score in our comparison, but produced the results under the same conditions by using the open-source code.}. We ran three additional baseline models in which the prediction stage is the same: (1) `No Context' uses no context summary; (2) `LSTM-Used Context Summary without Attention' uses the context summary of bidirectional LSTM without an attention mechanism, and (3) `LSTM-Used Content-Aware Attention' uses context summary of bidirectional LSTM after content-aware attention is applied to histories, as in previous approaches.

\subsection{Results}
We conducted an experiment to compare the proposed models with the baseline models in the SLU accuracy (Table  \ref{table:result}). All of the proposed models achieved significant improvements compared to all the baseline models.

We conducted an experiment to identify details of how possible combinations of the proposed methods affect the SLU accuracy (Table \ref{table:speaker_result}). In addition to the combinations of the proposed methods, we tested another content-and-time-aware attention method (Content x Time) which computes attention values using both intent and distance at a time, and shares the values to compare with the proposed content-and-time-aware attention.

\section{Discussion}
In the first subsection, we analyze the effectiveness of the decay-function-free time-aware attention and decay-function-free content-and-time-aware attention by comparison with others. In the next subsection, we analyze the effectiveness of the proposed methods in their possible combinations. We also analyze the effectiveness of the use of a distance vector in the history representation under the various conditions. Finally, we analyze attention weights of the proposed models in a qualitative way to convince of the effectiveness of them.

\begin{table}
  \centering
  \begin{tabulary}{5.0\textwidth}{p{2.6cm}|c|c}
    \hline
    &\multicolumn{2}{c}{\textbf{F1 score}} \\
    \textbf{Attention Type} & \textbf{Sent.-Level} & \textbf{Role-Level} \\
    \hline
    \hline
    no attention & 70.49 & 70.43 \\
    \hline
    Content & 73.03 & 72.87 \\
    \textbf{Content} & 73.05 & 72.68 \\
    \hline
    Time & 75.58 & 75.58 \\
    \textbf{Time} & 75.95\(^{**}\) & \textbf{76.56}\(^{**}\) \\
    \hline
    Content + Time & 75.59 & 75.30 \\
    \textbf{Content} + Time & 75.83 & 75.96\(^{**}\) \\
    Content + \textbf{Time} & 75.94\(^{*}\) & 75.97\(^{**}\) \\ 
    \textbf{Content} + \textbf{Time} & 76.11\(^{**}\) & 76.14\(^{**}\) \\
    \hline
    Content x Time & 75.59 & 75.50 \\
    \textbf{Content x Time} & 75.63 & 75.64 \\
    \hline
  \end{tabulary}
  \caption{SLU accuracy of possible combinations of the proposed methods. ``no attention'': sum of all history vectors without calculating \(\alpha\), ``Content'' is content-aware attention (Decay-Function-Free Content-Aware Attention), ``Time'' is the proposed time-aware attention (Decay-Function-Free Time-Aware Attention), ``Content + Time'' is the proposed content-and-time-aware attention (Decay-Function-Free Content-and-Time-Aware Attention), ``Content x Time'' is variant content-and-time-aware attention (Decay-Function-Free Inseparate Content-and-Time-Aware Attention). \(^{*}\): \(p < 0.05\); \(^{**}\), \(p < 0.01\) compared to the same attention without speaker indicator. An attention type in bold is the speaker-involved part.}
  \label{table:speaker_result}
\end{table}

We also conducted an experiment to show the effectiveness of the use of a distance vector in the history representation under the same condition as in the role-level attention (Table \ref{table:substance_result}). Although we propose to use both intent and distance by concatenating them as a history representation, intent can be used alone; this approach is more intuitive than using both intent and distance.

\begin{table}
  \centering
  \begin{tabulary}{5.0\textwidth}{p{2.6cm}|c|c}
    \hline
    &\multicolumn{2}{c}{\textbf{F1 score}} \\
    \textbf{Attention Type} & \textbf{Intent only} & \textbf{Int. \& Dist.} \\
    \hline
    \hline
    no attention & 70.20 & 70.43 \\
    Content & 71.09 & 72.87\(^{**}\) \\
    \textbf{Content} & 71.26 & 72.68\(^{**}\) \\
    Time & 75.17 & 75.58\(^{**}\) \\
    \textbf{Time} & 75.11 & \textbf{76.56}\(^{**}\) \\
    Content + Time & 75.04 & 75.30\(^{**}\) \\
    \textbf{Content} + Time & 75.62 & 75.96\(^{*}\) \\
    Content + \textbf{Time} & 75.13 & 75.97\(^{**}\) \\ 
    \textbf{Content} + \textbf{Time} & 75.67 & 76.14\(^{**}\) \\
    Content x Time & 75.08 & 75.50\(^{**}\) \\
    \textbf{Content x Time} & 75.03 & 75.64\(^{**}\) \\
    \hline
  \end{tabulary}
  \caption{SLU accuracy of possible combinations of the proposed methods in role-level attention with different history representations. Int. used intent vector; Dist. used distance vector. \(^{*}\): \(p < 0.05\); \(^{**}\), \(p < 0.01\) compared to using intent only. Other codes are as in Table \ref{table:speaker_result}.}
  \label{table:substance_result}
\end{table}

\subsection{Comparison with Baseline Models}
In Table \ref{table:result}, Decay-Function-Free Time-Aware Attention and Decay-Function-Free Content-and-Time-Aware Attention achieved significantly higher F1 scores that all baseline models. Especially, the role-level Decay-Function-Free Time-Aware Attention with speaker indicator achieved an F1 score of 76.56 \% (row 11), which is a state-of-the-art SLU accuracy.

\subsection{Detailed Analysis on Proposed Methods}
The proposed methods had good SLU accuracy (Table \ref{table:speaker_result}). Every time-aware attention with and without speaker indicator (rows 4 to 11) improved the F1 score compared to the content-aware attention with and without speaker indicator (rows 2 and 3) and to no attention (row 1). This result means that the proposed time-aware attention was effective to improve the SLU accuracy. Any of the content-and-time-aware attention with or without speaker indicator (rows 6 to 11) did not improve the F1 score compared to the time-aware attention with and without speaker indicator (rows 4 and 5). This result means that incorporating content could not make further improvement of the accuracy. Also, without speaker indicator, all the time-aware attention (rows 4, 6 and 10) achieved similar F1 scores.

\begin{figure*}[t]
  \centering
  \noindent
  \includegraphics[width=\linewidth]{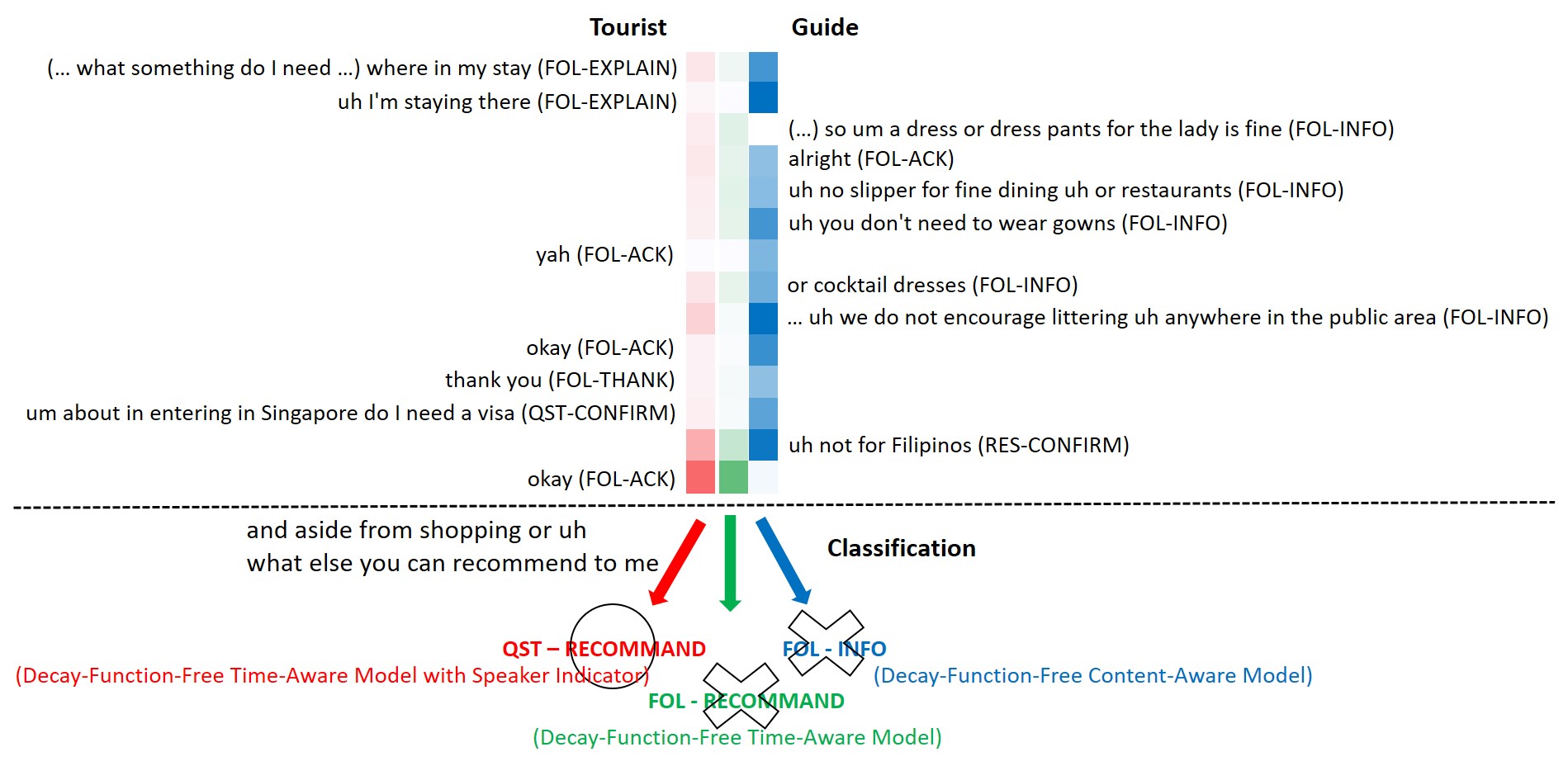}
  \caption{The visualization of the attention weights of the proposed models and the baseline content-aware model. Color gradient indicates intensity given a single datum after training. The color gradient at the left side indicates attention intensities of Decay-Function-Free Time-Aware Model with Speaker Indicator, the color gradient in the center indicates attention intensities of Decay-Function-Free Time-Aware Model without Speaker Indicator, and the color gradient at the right side indicates attention intensities of Decay-Function-Free Content-Aware Model.}
  \label{fig:vis_weights}
\end{figure*}


Use of speaker indicator also showed tendencies. It did not significantly improve the SLU accuracy of Decay-Function-Free Content-Aware Attention (rows 2 and 3) or Decay-Function-Free Inseparate Content-and-Time-Aware Attention (rows 10 and 11), but did improve the accuracy of the proposed models, Decay-Function-Free Time-Aware Attention (rows 4 and 5) and Decay-Function-Free Content-and-Time-Aware Attention (rows 6 to 9). Decay-Function-Free Content-and-Time-Aware Attention with speaker indicator (rows 7 to 9) were more accurate than Decay-Function-Free Inseparate Content-and-Time-Aware Attention with speaker indicator (row 11). This result means that using speaker indicator, separation of content and time improved the accuracy. The improvement in the role-level tended to be greater than that in the sentence-level. The improvement was greatest when speaker indicator was involved in the proposed role-level Decay-Function-Free Time-Aware Attention (row 5).

\subsection{Effectiveness of Use of Distance in History Representation}
In all models, the use of both intent and distance vectors significantly achieved higher F1 than the use of an intent vector only (Table \ref{table:substance_result}). The results indicate that distance embeddings are helpful both for attention and for the history representation. Decay-Function-Free Time-Aware Attention achieved the biggest improvement (row 5) among all the models.

\subsection{Qualitative Analysis}
To assess whether the proposed time-aware attention and speaker indicator can learn a time-decay tendency of the history effectively, we inspected the weights trained in Decay-Function-Free Time-Aware Attention with and without the speaker indicator. We also inspected Decay-Function-Free Content-Aware Attention to compare with them. We observed (Figure \ref{fig:vis_weights}) that the weights of the proposed models were trained well compared to Decay-Function-Free Content-Aware Attention. The proposed time-aware attention with/without speaker indicator tended to pay attention to recent histories, whereas the content-aware attention does not. As a result, Decay-Function-Free Time-Aware Attention with speaker indicator could generate the true label, \textit{QST-RECOMMEND}, by avoiding noisy contextual information like ``uh I'm staying there (FOL-EXPLAIN)'' or ``... uh we do not encourage littering uh anywhere in the public area (FOL-INFO)''.

\section{Conclusion}
In this paper, we propose decay-function-free time-aware attention models for SLU. These models summarize contextual information by taking advantage of temporal information without a manual time-decay function. We also propose a current-speaker detector that identifies the current speaker. In experiments on the DSTC 4 benchmark dataset, the proposed models achieved a state-of-the-art SLU accuracy. Detailed analysis of effectiveness of the proposed methods demonstrated that the proposed methods increase the accuracy of SLU individually. \\

\section*{Acknowledgments}
We would like to thank the reviewers for their insightful and constructive comments on this paper.

\bibliography{naaclhlt2019_mine}
\bibliographystyle{acl_natbib}

\end{document}